\documentclass[runningheads]{llncs}
\usepackage{graphicx}
\usepackage{algorithm}
\usepackage[noend]{algpseudocode}
\usepackage{multirow}
\usepackage{cite}
\usepackage{breqn}
\usepackage{amssymb}
\algrenewcommand{\algorithmiccomment}[1]{\hskip1em$\#$ #1}
\usepackage{url} 
\usepackage[colorlinks=true,linkcolor=blue,urlcolor=blue,citecolor=blue,anchorcolor=blue]{hyperref}

\begin{document}
\title{Solution and Fitness Evolution (SAFE): Coevolving Solutions and Their Objective Functions}
\titlerunning{Solution and Fitness Evolution (SAFE)}
%
\author{Moshe Sipper\inst{1,2} \and
        Jason H. Moore\inst{1}\and
        Ryan J. Urbanowicz\inst{1}
        \thanks{\textcopyright \hspace{2pt} Springer Nature Switzerland AG 2019. L. Sekanina et al. (Eds.): EuroGP 2019, LNCS 11451, pp. 1–16, 2019.
        The final authenticated version is available online at \url{https://doi.org/10.1007/978-3-030-16670-0_10}.}
        }
%
\institute{Institute for Biomedical Informatics, University of Pennsylvania, Philadelphia, PA 19104-6021, USA \and
Department of Computer Science, Ben-Gurion University, Beer Sheva 84105, Israel \\
\email{\{sipper,jhmoore,ryanurb\}@upenn.edu}\\
}
\maketitle              
\begin{abstract} \sloppy
We recently highlighted a fundamental problem recognized to confound algorithmic optimization, namely, \textit{conflating} the objective with the objective function. Even when the former is well defined, the latter may not be obvious, e.g., in learning a strategy to navigate a maze to find a goal (objective), an effective objective function to \textit{evaluate} strategies may not be a simple function of the distance to the objective. We proposed to automate the means by which a good objective function may be discovered---a proposal reified herein. We present \textbf{S}olution \textbf{A}nd \textbf{F}itness \textbf{E}volution (\textbf{SAFE}), a \textit{commensalistic} coevolutionary algorithm that maintains two coevolving populations: a population of candidate solutions and a population of candidate objective functions. As proof of principle of this concept, we show that SAFE successfully evolves not only solutions within a robotic maze domain, but also the objective functions needed to measure solution quality during evolution.

\end{abstract}

\section{Objective $\neq$ Objective Function}
\begin{figure}
\centering
\begin{tabular}{c@{\hskip 20px}c}
\includegraphics[width=0.4\textwidth]{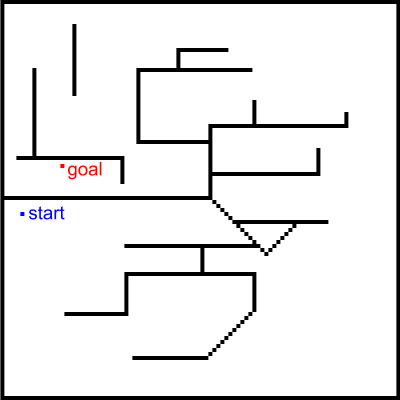} &
\includegraphics[width=0.4\textwidth]{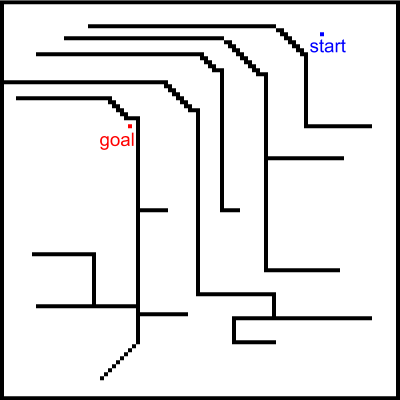} \\
\texttt{maze1} & \texttt{maze2}
\end{tabular}
\caption{In these maze problems a robot begins at the start square and must make its way to the goal square (objective).} \label{fig:maze}
\end{figure}

The goal of any evolutionary algorithm (EA) is to solve a problem, i.e., obtain a specified \textit{objective}. This invariably entails defining an \textit{objective function} (e.g., fitness function), which is the function we want to minimize or maximize.  
In a recent paper, we targeted a fundamental 
problem that one might face in practice.  Specifically, while the objective may be known and well defined, the objective function may be deceptive, and one might easily conflate the objective with the objective function \cite{Sipper2018}. Consider the mazes in Figure~\ref{fig:maze}, wherein the challenge is to evolve a robotic controller (i.e., a model that determines movement given the robot's current state) such that the robot, when placed in the start position, is able to make its way to the goal. The controller defines the \textit{behavior} (i.e., decides the direction of movement) of the robot when it is in a given \textit{state} (position in the maze, distances to obstacles). The set of movement decisions over a fixed number of time steps defines the robot's \textit{path}, and the robot's \textit{endpoint} is its position at the final time step. 

\begin{figure}
\centering
\begin{tabular}{c@{\hskip 20px}c}
\includegraphics[width=0.4\textwidth]{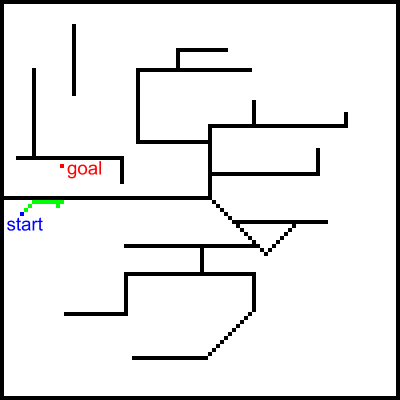} &
\includegraphics[width=0.4\textwidth]{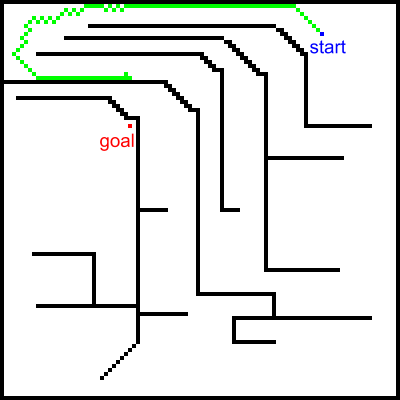} \\
\texttt{maze1} & \texttt{maze2}
\end{tabular}
\caption{Path (green) of a robot evolved by a standard evolutionary algorithm with fitness measured as distance-to-goal, evidencing how conflating the objective with the objective function leads to a non-optimal solution.} \label{fig:maze-ea}
\end{figure}

It seems intuitive that the fitness $f$ of a given robotic controller be defined as a function of the distance from the robot to the objective, as done, e.g., by \cite{Lehman2008}. However, reaching the objective may be difficult since the robot is faced with a deceptive landscape, where higher fitness (i.e., being reasonably close to the goal) may not imply that the robot is ``almost there''. It is quite easy for the robot to attain a fairly good fitness value, yet be stuck behind a wall in a local optimum---quite far from the objective in terms of the path needed to be taken. Indeed, our experiments with such a fitness-based evolutionary algorithm (Section~\ref{sec:results}) produced the expected failure, demonstrated in Figure~\ref{fig:maze-ea}.

Reaching the global optimum implies the acceptance of \textit{reduced} fitness over the course of searching for an optimal controller. This is at odds with the workings of an evolutionary algorithm, which in practice is driven to optimize the objective function rather than find the best way to reach the objective. Another way to frame this problem is that while an optimal solution may be representable by the controller, it may not be learnable given this simple objective function \cite{domingos2012few}. For EAs, learnability translates to \textit{evolvability} \cite{wagner1996perspective}. 

One solution to this conflation problem was offered by \cite{Lehman2008} in the form of novelty search, which \textit{ignores} the objective and searches for novelty (using a novelty metric that requires careful consideration, e.g., robot endpoint novelty). However, novelty for the sake of novelty alone lacks incentive for solutions that reach and stay at the objective. In \cite{Sipper2018} we offered an alternative view, namely, that the problem may lie with our ignorance of the \textit{correct} objective function. We proposed that rather than exclusively seek out novelty or rely on a fixed objective-based metric to evaluate both the ``journey'' and the ``destination'' it might be more effective to separate the optimization of the solution from the optimization of the objective function. For example, just because the robot's objective is to reach the goal does not necessarily imply that the objective function is distance-to-goal. 

Thus, we reasoned, we are now left with the task of finding this better objective function, either manually or---perhaps more intriguingly---through some automated means; after all, if we are searching for a good objective function, why not employ a search algorithm? We concluded in \cite{Sipper2018} with a final speculation, namely, that coevolution \cite{zaritsky2004,Pena:2001} might offer the key to simultaneously explore solution optimality and objective-function optimality.

In this paper we flesh out our previous speculations and offer preliminary evidence that solutions can be coevolved along with the objective functions that evaluate them. Our newly proposed algorithm is dubbed \textbf{SAFE}, for \textbf{S}olution \textbf{A}nd \textbf{F}itness \textbf{E}volution. 

This paper describes an exploratory study of SAFE. We show that where a standard, fitness-based evolutionary algorithm fails, SAFE succeeds, on a par with (or slighly better) than novelty search.

We examine four algorithms in this work---(1) a standard genetic algorithm, (2) novelty search, (3) SAFE, and (4) random search---for their ability to solve maze problems. After recounting related work in Section~\ref{sec:previous},
we present two key ideas, novelty search (Section~\ref{sec:novelty}) and coevolution (Section~\ref{sec:coevolution}). We continue with a delineation of the robot simulator (Section~\ref{sec:robot}). We then present SAFE (Section~\ref{sec:safe}) followed by results involving maze-solving robots (Section~\ref{sec:results}). We end with a discussion and concluding remarks (Section~\ref{sec:discussion}).

\section{Related Work}
\label{sec:previous}

We believe that the idea presented herein---treating the objective function as an evolving entity---has not been previously examined as such. There are a number of related, but distinct, research directions summarized below.

\textit{Fitness approximation} explores the ways one can estimate the fitness function by constructing an approximate model. This can be quite useful when an explicit fitness function does not exist, or the evaluation of the fitness is computationally expensive. An excellent review on the topic by \cite{jin2005comprehensive} divides fitness-approximation approaches into three categories: 1) \textit{Problem approximation} tries to replace the original statement of the problem by one which is approximately the same but easier to solve (e.g., evaluate a turbine blade using computational fluid dynamics---CFD---simulations rather than in a wind tunnel); 2) \textit{functional approximation} constructs an alternate expression for the objective function, e.g., instead of a CFD simulation of a blade define an explicit mathematical
model (see also work on surrogate fitness functions \cite{buche2005accelerating,Brownlee2010}); 3) \textit{evolutionary approximation}, including fitness inheritance---wherein offspring fitness is estimated from parent fitness, and fitness imitation---where individuals are clustered into groups and only a group representative is evaluated whereupon it serves as a basis for estimating the fitness of the the others in the cluster. Interestingly, coevolution has been used in fitness approximation \cite{schmidt2008coevolution}.

The objective function conflation problem is also distinct from the topic of \textit{dynamic fitness}, wherein the problem itself (along with the global optimum and/or the fitness landscape) changes during run-time \cite{grefenstette1999evolvability}. The prediction of weather patterns is in line with this challenge. Further, \textit{multi-objective optimization}, where there is more than one objective to optimize at a time, is also distinct \cite{deb2000fast} given that it is typically up to the user to define a multi-objective fitness function. Multi-objective Pareto front methods are perhaps closest to the concept we propose in that they seek not to make fixed assumptions about what makes a `good' solution/objective \cite{deb2000fast}. However, a Pareto front does not separate the objective from the objective function, and ultimately leaves the challenge of picking the `optimal' solution up to the user, traditionally from a set of candidate `non-dominated' solutions at the front. Ultimately, the challenges of dynamic fitness and multi-objective optimization are each vulnerable to the conflation of objective and objective function---and might benefit from a SAFE-inspired approach. 

More generally, the topic of \textit{open-ended evolution} comes to mind. This type of evolution occurs in nature, admitting no externally imposed fitness criterion, but rather an implicit, emergent, dynamical one (that could arguably be summed up as survivability) \cite{sipper1997if, sipper02mn,banzhaf2016defining}. 
Indeed, the original novelty-search paper began with: ``This paper establishes a link between the challenge of solving
highly ambitious problems in machine learning and the
goal of reproducing the dynamics of open-ended evolution in
artificial life.'' \cite{Lehman2008}
While open-ended evolution is a popular topic in the field of artificial life, we prefer to keep the focus herein on computational problem solving.

\section{Novelty Search}
\label{sec:novelty}

Novelty search was introduced by \cite{Lehman2008} and applied to the examination of maze problems where robot controllers were sought as candidate solutions. Recall that a controller defines the behavior of a robot, yielding a path through the maze concluding in an endpoint position. The key idea was that instead of rewarding endpoint closeness to objective, individual solutions would be considered valuable if their endpoint diverged from prior solution endpoints (i.e., the system was driven to identify solutions that led the robot to new parts of the maze, regardless of objective closeness). 

The novelty metric employed should be carefully defined based on the problem at hand. In \cite{Lehman2008}, the novelty metric was defined as the Euclidean distance between the endpoints of two individuals. Note that this is essentially a \textit{phenotypic}, or outcome-based, novelty measure. Differently, a \textit{genotypic} novelty metric would examine differences between the architecture of the solutions themselves (e.g., the robot controllers). We will revisit this in Section~\ref{sec:safe}, where we present a genotypic novelty metric for SAFE objective functions. 

In later work, novelty was combined with other objectives (e.g., performance). For example, in \cite{lehman2011evolving}, ``an existing creature evolution platform is extended with multi-objective search that balances drives for both novelty and performance''. However, this does not involve an evolving objective function, but rather, ``The suggested approach is to provide evolution with both a novelty objective \ldots and a local competition objective''. To wit, the objectives are provided by the user, as opposed to SAFE, wherein they are evolved. Appendix F of \cite{stanley2018art} includes a comprehensive up-to-date list of references to research involving novelty search.

Effectively, novelty search is identical to a standard evolutionary algorithm, with the fitness function replaced by the novelty metric. We seek novel behaviors rather than the objective, hoping that the former will ultimately lead to the latter.
The endpoint of a candidate solution is compared to its cohorts in the current population and to an archive of past individuals whose behavioral endpoints were highly novel when they emerged. The novelty score is the average of the distances to the $k$ nearest neighbors ($k$ was set to 15 both by \cite{Lehman2008} and by us; see Table~\ref{tab:params} in Section~\ref{sec:results} for a summary of parameters).

For future problem flexibility, we implemented the novelty archive differently than \cite{Lehman2008}. There, an individual robot's endpoint entered the archive if its novelty score was above a certain threshold (which was adjusted dynamically). We chose to implement a fixed-size archive, with new points entering the archive until it fills up. Once full, a newly found point that exceeds the archive's current minimum will replace that minimum; otherwise it will not enter the archive. Thus, the archive saves the most novel points (at the time of their emergence). Our choice of a different archive implementation stems from our initial forays into other problems (e.g., function optimization) ---to be reported in the future---where no threshold value was readily apparent.

To summarize, a robot (controller's) novelty score is computed by having the robot wander the maze for the allotted number of steps (300---see Table~\ref{tab:params}). Then, its endpoint is compared to all endpoints of its cohorts in the current generation \textit{and} to all endpoints in the archive. The final score is then the average of the $k=15$ nearest neighbors.

\section{Coevolution}
\label{sec:coevolution}

Coevolution refers to the simultaneous evolution of two or more species with coupled fitness \cite{Pena:2001}. Strongly related to the concept of symbiosis, coevolution can be mutualistic, parasitic, or commensalistic \cite{wiki:Symbiosis}:
1) In mutualism, different species exist in a relationship in which each individual (fitness) benefits from the activity of the other; 2) in parasitism, an organism lives on or in another organism, causing it  harm; and 3) in commensalism, members of one species gain benefits while those of the other species neither benefit nor are harmed.
Interestingly, though the idea of coevolution originates (at least) with Darwin---who spoke of ``coadaptations of organic beings to each other'' \cite{Darwin:1859}---it is arguably somewhat less pervasive in the field of evolutionary computation than one might expect.

A cooperative (mutualistic) coevolutionary algorithm involves a number of independently evolving species, which come together to obtain problem solutions. The fitness of an individual depends on its ability to collaborate with individuals from other species \cite{zaritsky2004,Pena:2001,Potter:2000,Dick:2014}. 

In a competitive (parasitic) coevolutionary algorithm the fitness of an individual is based on direct competition with individuals of other species, which in turn evolve separately in their own populations. Increased fitness of one of the species implies a reduction in the fitness of the other species \cite{Hillis:1990}. 

We will discuss the particulars of coevolution---specifically, commensalistic co\-evolution---in Section~\ref{sec:safe}, when we present the SAFE algorithm. To our knowledge, this is the first introduction of a commensalistic coevolutionary algorithm.

\section{The Robot and the Maze}
\label{sec:robot}

We implemented a robotic simulator, similar to the one used by \cite{Lehman2008}, involving a robot moving in a two-dimensional environment that contains walls, with a controller dictating its actions, based on inputs from several sensors. The major difference from \cite{Lehman2008} is that we did not use NEAT---NeuroEvolution of Augmenting Topologies---for the controller, as we wanted full access to all code aspects, and we wanted to make our controller as interpretable as possible.   
 
The maze environment is a two-dimensional grid of size $100 \times 100$ cells, where each cell can either be empty or (part of a) wall (refer to Figure~\ref{fig:maze}). Two cells are designated as the start and goal. The robot can move horizontally or vertically, but not diagonally. The geometry is taxicab (Manhattan), i.e., the distance between two points is the sum of the absolute differences of their Cartesian coordinates.

\begin{figure}
\centering
\includegraphics[width=0.6\textwidth]{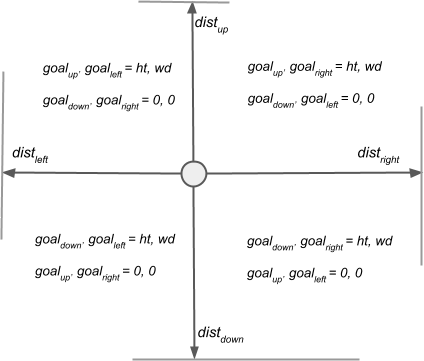}
\caption{The robot's 4 $\mathit{dist}$ sensors indicate distance to nearest obstacle (wall or boundary). The 4 $\mathit{goal}$ sensors indicate direction of goal: 2 are always set to $\mathit{ht}$ and $\mathit{wd}$ (maze height and width), and 2 are set to $0$.} \label{fig:sensors}
\end{figure}

The robot has 8 sensors (Figure~\ref{fig:sensors}): 
\begin{itemize}
\item 
$\mathit{dist_{up}}$, 
$\mathit{dist_{down}}$, 
$\mathit{dist_{left}}$, 
$\mathit{dist_{right}}$: 
distances to the obstacle (or boundary) in the upward, downward, leftward, and rightward direction, respectively.

\item 
$\mathit{goal_{up}}$, 
$\mathit{goal_{down}}$, 
$\mathit{goal_{left}}$, 
$\mathit{goal_{right}}$: 
indicators of whether the goal is towards the upward, downward, leftward, or rightward direction, respectively.
\end{itemize}
The $\mathit{dist}$ values are computed by examining the robot's current position with respect to the nearest obstacle or boundary in the appropriate direction. 
The $\mathit{goal}$ values are computed with respect to the robot's current position (the robot can sense the goal ``beacon'' through walls). If the goal is in the northeast quadrant of the robot, $\mathit{goal_{up}}$ and $\mathit{goal_{right}}$ are set to the maze height and width, respectively, while the other two $\mathit{goal}$ sensors are set to $0$. Similarly, the sensors are set when the goal is northwest, southeast, or southwest of the robot (Figure~\ref{fig:sensors}). 

The robot moves for an allotted number of steps (set to 300, see Table~\ref{tab:params}). Movement is controlled by two variables, horizontal -- $h$ and vertical -- $v$. The robot moves one cell in the horizontal position and one cell in the vertical position at every step, as follows:
\begin{enumerate}
    \item Compute robot's 8 sensor values.
    \item Set $h$ and $v$ to:
         \begin{dmath*}
            h=
            p_1\mathit{dist_{right}} + 
            p_2\mathit{goal_{right}} + 
            p_3\mathit{dist_{left}} +
            p_4\mathit{goal_{left}} + 
            p_5\mathit{dist_{down}} + 
            p_6\mathit{goal_{down}} + 
            p_7\mathit{dist_{up}} + 
            p_8\mathit{goal_{up}} 
         \end{dmath*}

         \begin{dmath*}
            v=
            p_9\mathit{dist_{right}}+ 
            p_{10}\mathit{goal_{right}} + 
            p_{11}\mathit{dist_{left}} +
            p_{12}\mathit{goal_{left}} + 
            p_{13}\mathit{dist_{down}} + 
            p_{14}\mathit{goal_{down}} + 
            p_{15}\mathit{dist_{up}} + 
            p_{16}\mathit{goal_{up}} 
         \end{dmath*}
    
    \item If $h \geq 0$ ($h < 0$) then move right (left), provided adjacent cell is neither boundary nor wall (otherwise don't move).
    
    \item If $v \geq 0$ ($v < 0$) then move down (up), provided adjacent cell is neither boundary nor wall (otherwise don't move).
\end{enumerate}

The solution search space is engendered by the vector $[p_1, \ldots, p_{16}] \in \mathbb{R} \cap [-1,1]$, where each set of $[p_1, \ldots, p_{16}]$ values constitutes a robot's control vector, which determines its behavior. In this domain, a successful vector will drive the robot to the goal. The algorithms we examine below all aim to find a successful control vector.

\section{SAFE: Solution And Fitness Evolution}
\label{sec:safe}

\begin{figure}
\centering
\includegraphics[width=0.97\textwidth]{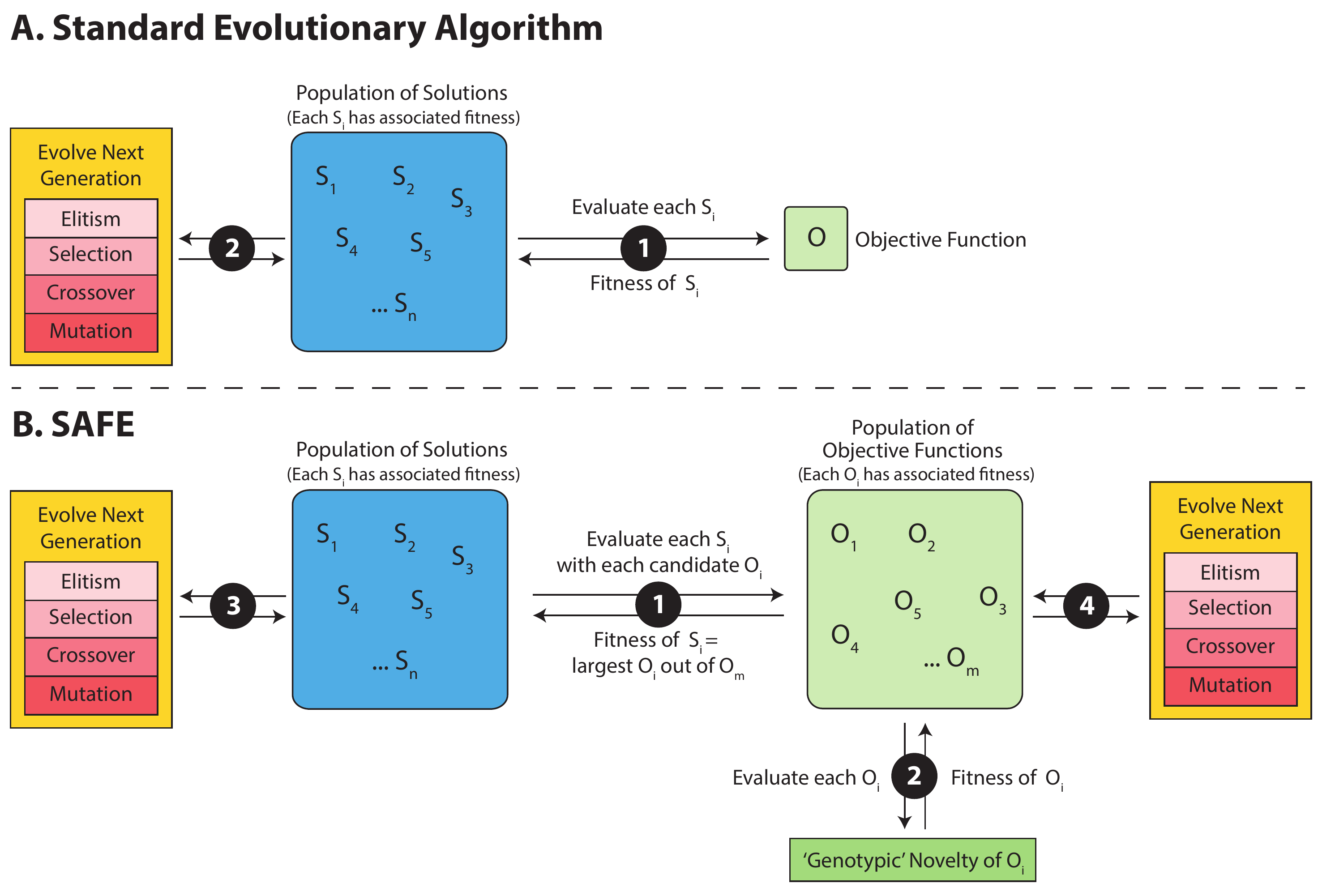} 
\caption{A single generation of a standard evolutionary algorithm vs. a single generation of SAFE. The numbered circles identify sequential steps in the respective algorithms.} \label{fig:safe}
\end{figure}

SAFE is a coevolutionary algorithm that maintains two coevolving populations: a population of candidate solutions and a population of candidate objective functions (see Figure~\ref{fig:safe}).
The evolution of each population is identical to a standard, single-population evolutionary algorithm---except where fitness computation is concerned.
Below we describe the various components of the system: population composition, initialization, selection, elitism, crossover, mutation, and fitness computation.

\textbf{Populations.}
An individual in the solutions population is a list of 16 real values, each in the range $[-1,1]$. These values are the robot parameters, $p_i$, described in Section~\ref{sec:robot}.

An individual in the objective-functions population is a list of 2 real values $[a,b]$, each in the range $[0,1]$, whose usage is described below. Population sizes and other parameters are given in Table~\ref{tab:params} (Section~\ref{sec:results}).

\textbf{Initialization.}
For every evolutionary run: both populations are initialized to random (fixed-length) lists, wherein each component value is in the  appropriate range.

\textbf{Selection.} 
Tournament selection with tournament size 5, i.e., choose 5 individuals at random from the population and return the individual with the best fitness as the selected one.

\textbf{Elitism.} The 2 individuals with the highest fitness in a generation are copied (``cloned'') into the next generation unchanged.

\textbf{Crossover.} Standard single-point crossover, i.e., select a random crossover point and swap two parent genomes beyond this point to create two offspring. The crossover rate is the probability with which crossover between two selected parents occurs. (Note that for an objective-function individual, which comprises two values, if crossover occurs, it will always be, ipso facto, at the same position.)

\textbf{Mutation.} 
Mutation is done with probability 0.4 (per individual in the population) by selecting a random gene (of the 16 or 2, respectively) and replacing it with a new random value in the appropriate range. 

\textbf{Fitness.}
Fitness computation is where SAFE dynamics come into play. 

In a standard evolutionary algorithm (which is one of the four algorithms we test below) each solution individual is assigned fitness based on a fixed objective function that computes a value inversely proportional to the distance to goal of the robot's endpoint, i.e., $1/\mathit{distToGoal}$ (Figure~\ref{fig:safe}A). 

With novelty search, an individual's fitness value is replaced with its novelty score, as described in Section~\ref{sec:novelty}.

In SAFE, each solution individual, $S_i$, $i \in \{1,\ldots,n\}$ is scored by every candidate objective-function individual, $O_j$, in the current population, $j \in \{1,\ldots,m\}$ (Figure~\ref{fig:safe}B). In this preliminary investigation, candidate SAFE objective functions were allowed to incorporate both `distance to goal' as well as novelty in order to calculate solution fitness. The best (highest) of these objective function scores is then assigned to the individual solution as its fitness value (Algorithm~\ref{alg:sol-fit}). Note that most of the computational cost goes into simulating the robot and computing novelty scores---which is done only once per individual solution.

\begin{algorithm}
\caption{Compute fitness values of solutions population}\label{alg:sol-fit}
\begin{algorithmic}[1]
\State $n$ $\gets$ size of solutions population
\State $m$ $\gets$ size of objective-functions population
\For{$i \gets 1$ to $n$} 
\State simulate robot with $S_i$, and derive $\mathit{endPosition_i}$, $\mathit{distToGoal_i}$
\EndFor

\For{$i \gets 1$ to $n$} 
\State compute $\mathit{noveltyScore_i}$, based on $\mathit{endPosition}$ and $\mathit{archive}$
\EndFor

\For{$i \gets 1$ to $n$} 
\For{$j \gets 1$ to $m$} 
\State $f_j \gets \mathit{O_j(S_i)}$ \Comment{$\mathit{O_j}$ uses $\mathit{distToGoal}$ \& $\mathit{noveltyScore}$ (Equation~\ref{eq:obj})}
\EndFor
\State $\mathit{solutionFitness_i} \gets \max f_j$
\EndFor
\end{algorithmic}
\end{algorithm}

As noted above, an objective-function individual is a pair $[a,b]$; specifically, $a$ determines the influence of `distance to goal' and $b$ determines the influence of `phenotypic solution novelty'. $O_j(S_i)$ is the fitness score that objective function  $O_j$ assigns to solution $S_i$, given as: 
\begin{equation}\label{eq:obj}
O_j(S_i) = a \times \frac{1}{\mathit{distToGoal_i}} +
      b \times \mathit{noveltyScore_i} \, ,
\end{equation}
where 
$\mathit{distToGoal_i}$ is the distance to goal of robotic controller (solution) $i$'s endpoint, and $\mathit{noveltyScore_i}$ is the novelty score of solution $i$. Rather than use one or the other, we let evolution discover an effective mix of the two objective function components (as opposed, e.g., to \cite{Cuccu2011}, wherein a  weighted combination of fitness and novelty was used with pre-determined weights).

As for the objective-functions population, determining the quality of an evolving objective function places us in uncharted waters. Such an individual is not a solution to a problem, but rather the ``guide''---or ``path''---to a solution. As such, it is not clear what comprises a good measure of success. 

We experimented with various forms of mutualistic coevolution, where the fitness of an objective function depends on its ability to ascribe fitness to solutions in the solutions population, such that better solutions (i.e., closer to the objective) receive higher fitness values, and worse solutions receive lower fitness values. In particular, we considered evaluating objective functions based on the correlation (i.e., Pearson's or Spearman's) between the solution fitness scores it generated and distance to the actual objective. While seemingly intuitive, this approach yielded unsatisfactory results, finding that this implementation ended up reinforcing the same local minima problem encountered when using a traditional evolutionary algorithm. Despite this failure, we plan to continue investigating the possibility of a mutualistic coevolutionary approach, where the fitness of objective functions is dependent in some way on the population of candidate solutions. 

In the working version of SAFE presented here, we turned to a commensalistic coevolution strategy, where the objective functions' fitness does not depend on the population of solutions. Instead, it relies on \textit{genotypic} novelty, based on the objective-function individual's two-valued genome, $[a,b]$. The distance between two objective functions---$[a_1,b_1]$, $[a_2,b_2]$---is simply the Euclidean distance of their genomes, given as: $\sqrt{(a_1-a_2)^2 + (b_1-b_2)^2}$. Note the contrast between genotypic novelty here and phenotypic (i.e., outcome-based) novelty used to evaluate solutions (both by novelty search and by SAFE).

Each generation, every candidate objective function is compared to its cohorts in the current population of objective functions and to an archive of past individuals whose behaviors were highly novel when they emerged. The novelty score is the average of the distances to the $k$ ($=15$) nearest neighbors, and is used in computing objective-function fitness (Algorithm~\ref{alg:obj-fit}). 

\begin{algorithm}
\caption{Compute fitness values of objective-functions population}\label{alg:obj-fit}
\begin{algorithmic}[1]
\State $m$ $\gets$ size of objective-functions population
\For{$i \gets 1$ to $m$} 
\State compute $\mathit{noveltyScoreObj_i}$
\State $\mathit{objectiveFitness_i} \gets \mathit{noveltyScoreObj_i}$ 
\EndFor
\end{algorithmic}
\end{algorithm}

To clarify, we recap SAFE's double use of the idea of novelty. First, solutions evolve not via pure goal-directedness or through pure phenotypic novelty, but rather through a combination of the two. The exact nature of this combination is determined by the evolving objective functions, which themselves use novelty (albeit genotypic) to explore the objective-function search space.

The code for SAFE and all the experiments carried out in this paper is available at \url{https://github.com/EpistasisLab/}.

\section{Results}
\label{sec:results}

To test our new framework, we performed eight sets of experiments, using four algorithms, each set to search for robotic controllers that solve the two mazes of Figure~\ref{fig:maze}:
\begin{enumerate}
    \item \sloppy Standard evolutionary algorithm (fitness is $1/\mathit{distToGoal}$).
    \item Novelty search.
    \item SAFE.
    \item Random search.
\end{enumerate}
Table~\ref{tab:params} summarizes the run parameters.

\begin{table}
\centering
\caption{Algorithm parameters. Unless stated, a relevant parameter not shown is that listed under `Standard EA' (e.g., the generation count for novelty search is that of the standard EA).}\label{tab:params}
\begin{tabular}{|l|l|}
\hline
Description & Value \\ 

\hline 
\multicolumn{1}{|c|}{Standard EA} & \\ \hline
Number of evolutionary runs & 500  \\
Maximal number of generations & 500 \\
Size of solutions population & 200 \\
Type of selection & Tournament \\
Tournament size & 5 \\
Type of crossover & single-point \\
Crossover rate & 0.8 \\
Probability of mutation (solutions) & 0.4 \\
Number of top individuals copied (elitism) & 2 \\
Maximal no. steps taken by robot & 300 \\
Stop if distance to goal $\leq$ & 2 \\

\hline
\multicolumn{1}{|c|}{Novelty search} & \\ \hline
Average over $k$ nearest neighbors, $k=$ & 15 \\
Archive size & 1000 \\

\hline
\multicolumn{1}{|c|}{SAFE} & \\ \hline
Size of objective-functions population & 200 \\
Probability of mutation (objective functions) & 0.4 \\

\hline
\multicolumn{1}{|c|}{Random search} & \\ \hline
Number of random individuals drawn & 100,000 \\

\hline
\end{tabular}
\end{table}

Random search serves as a yardstick to ensure evolution is tackling a non-trivial task. It is done by drawing 100,000 solutions at random and outputting the best one. 100,000 equals maximal number of generations $\times$ population size, which are used by the previous algorithms. This ensures a fair comparison in terms of resources. 

\begin{figure}
\centering
\begin{tabular}{c@{\hskip 20px}c}
\includegraphics[width=0.4\textwidth]{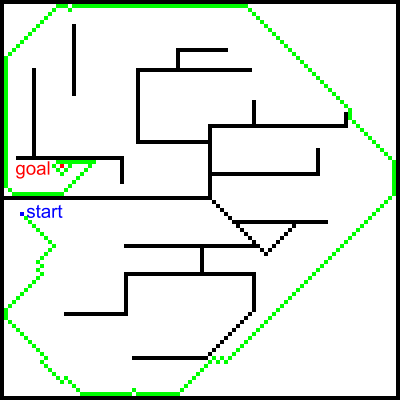} &
\includegraphics[width=0.4\textwidth]{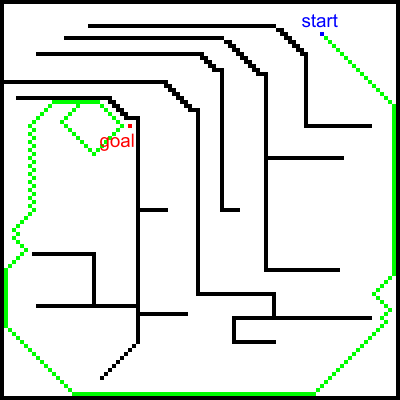} \\
\texttt{maze1} & \texttt{maze2}
\end{tabular}
\caption{Solutions to the maze problems, evolved by SAFE.} \label{fig:maze-solutions}
\end{figure}

The results are shown in Table~\ref{tab:maze}. A solution is deemed successful if the robot gets to within a distance of 2 from the goal. We note that SAFE is on a par with novelty search on \texttt{maze1}, and able to discover slightly more solutions for the harder \texttt{maze2}. SAFE also appeared to find a solution that reached the goal within slightly fewer generations, however this finding cannot be deemed significant. Figure~\ref{fig:maze-solutions} shows sample solutions found by SAFE (contrast this with the  standard evolutionary algorithm, which always gets stuck in a local minimum, as exemplified in Figure~\ref{fig:maze-ea}). 

\begin{table}
\centering
\caption{Results of 500 evolutionary runs per each algorithm. Success: number of successful runs, where the algorithm discovered a solution that gets the robot to a distance of 2 or less from the goal. Generations: average number of generations (standard deviation) to success.}\label{tab:maze}
\begin{tabular}{|r|c|c|c|}
\hline
Algorithm & Maze & Success & Generations \\
\hline
\multirow{2}{*}{Standard EA} & \texttt{maze1} & 0 & --- \\
                             & \texttt{maze2} & 0 & --- \\
\hline
\multirow{2}{*}{Novelty} & \texttt{maze1} & 322 & 145 (132) \\
                         & \texttt{maze2} & 10 & 249 (163) \\
\hline
\multirow{2}{*}{SAFE} & \texttt{maze1} & 328 & 141 (130) \\
                      & \texttt{maze2} & 17 &  227 (152)\\
\hline
\multirow{2}{*}{Random} & \texttt{maze1} & 9 & --- \\
                        & \texttt{maze2} & 0 & --- \\\hline
\end{tabular}
\end{table}

As explained in Section~\ref{sec:safe}, the evolving objective function is a real-valued tuple $[a,b]$, with $a$ being the distance-to-goal coefficient, and $b$ being the novelty coefficient. Examining all successful objective functions evolved by SAFE for \texttt{maze1} (those that led to a maze solution), we observed that the average of $a$ was 0.85 ($\mathit{sd}=0.16$) and the average of $b$ was 0.99 ($\mathit{sd}=0.01$); for \texttt{maze2}, average $a$ was 0.83 ($\mathit{sd}=0.18$) and average $b$ was 0.98 ($\mathit{sd}=0.02$).
Evolution is thus seen to strike a fairly even balance between being goal-oriented and seeking novelty, favoring the latter to a small extent. This motivated a final experiment, wherein we ran the standard evolutionary algorithm, but with the fitness function of SAFE (Equation~\ref{eq:obj}) and fixed $a=b=0.5$. The number of successful runs was 307 for \texttt{maze1} and 17 for \texttt{maze2}, on par with SAFE and novelty search. Of course, SAFE's task is more difficult, given that it is not handed the objective function but must evolve it. 

Recall from Section~\ref{sec:robot} that the robot's movement is controlled by two variables, horizontal -- $h$ and vertical -- $v$, 
each defined as a weighted sum of the eight sensor values:
$\mathit{dist_{right}}$,  
$\mathit{goal_{right}}$,  
$\mathit{dist_{left}}$, 
$\mathit{goal_{left}}$,  
$\mathit{dist_{down}}$,  
$\mathit{goal_{down}}$,  
$\mathit{dist_{up}}$, 
$\mathit{goal_{up}}$.
An evolved solution is a vector of 16 real values, 
$[p_1, \ldots, p_{16}]$, where $[p_1, \ldots, p_8]$ are the weights used by $h$, and $[p_9, \ldots, p_{16}]$ are the weights used by $v$.
Given that the model is fairly white box in nature (as opposed, e.g., to a neural network's black-box essence), we might attempt to examine the evolved solutions---whose summary is shown in Table~\ref{tab:evolved-params}. 
For \texttt{maze1} we note that $h$'s $\mathit{left}$ (absolute) sensor values are quite low, perhaps because the start position is close to the left wall, so there is no point in going left. Also, the $\mathit{right}$ values are high for both $h$ and $v$, so apparently right is a good direction. For \texttt{maze2} we observe that moving left seems to be favored, and---to a lesser degree---down. 

\begin{table}
\centering
\caption{Evolved robotic control parameters---$[p_1, \ldots, p_{16}]$---of successful solutions. Each value is the average (standard deviation) of all successful runs. Parameters are shown alongside the sensors they weight.} \label{tab:evolved-params}
\begin{tabular}{|c|l|c|c|}
\hline
p & sensor & \texttt{maze1} & \texttt{maze2} \\
\hline
\multicolumn{4}{|l|}{horizontal}  \\ \hline
$p_1$    & $\mathit{dist_{right}}$ & 0.68 (0.26)  & -0.27 (0.33) \\
$p_2$    & $\mathit{goal_{right}}$ & -0.56 (0.29) & 0.04 (0.38)  \\
$p_3$    & $\mathit{dist_{left}}$  & 0.11 (0.39)  & -0.82 (0.18) \\
$p_4$    & $\mathit{goal_{left}}$  & -0.04 (0.29) & 0.52 (0.4) \\
$p_5$    & $\mathit{dist_{down}}$  & -0.65 (0.25) & 0.58 (0.42) \\
$p_6$    & $\mathit{goal_{down}}$  & -0.73 (0.2)  & 0.38 (0.45) \\
$p_7$    & $\mathit{dist_{up}}$    & 0.45 (0.38)  & 0.11 (0.2) \\
$p_8$    & $\mathit{goal_{up}}$    & 0.33 (0.28)  & -0.37 (0.37) \\
\hline
\multicolumn{4}{|l|}{vertical}  \\ \hline
$p_9$    & $\mathit{dist_{right}}$ & 0.65 (0.23)  & 0.1 (0.1) \\
$p_{10}$ & $\mathit{goal_{right}}$ & 0.62 (0.31)  & -0.09 (0.45) \\
$p_{11}$ & $\mathit{dist_{left}}$  & -0.4 (0.46)  & 0.78 (0.15) \\ 
$p_{12}$ & $\mathit{goal_{left}}$  & -0.55 (0.31) & 0.54 (0.29) \\
$p_{13}$ & $\mathit{dist_{down}}$  & -0.24 (0.46) & 0.11 (0.17) \\
$p_{14}$ & $\mathit{goal_{down}}$  & 0.03 (0.38)  & -0.15 (0.5) \\
$p_{15}$ & $\mathit{dist_{up}}$    & -0.64 (0.34) & -0.21 (0.39) \\
$p_{16}$ & $\mathit{goal_{up}}$    & 0.41 (0.3)   & -0.15 (0.52) \\
\hline
\end{tabular}
\end{table}

\section{Discussion and Concluding Remarks}
\label{sec:discussion}

Aiming to confront the optimization conflation problem---where the objective is conflated with the objective function---we separated these two entities into two populations, and presented SAFE, a coevolutionary algorithm to evolve the two simultaneously. We showed that both the ``journey'' and the ``destination'' can be discovered together, i.e., the solution and the objective function needed to find it. 

Our aim herein has been to provide preliminary proof for a novel idea. As such, we did not perform extensive experiments on parameter combinations, although recent research suggests that parameters should not matter too much \cite{Sipper2018par}. 

Stanley \cite{stanley2018art} recently wrote that, ``While a good algorithm is sometimes one that performs well, sometimes a good algorithm is instead \textit{one that leads to other algorithms and new frontiers}.''
While ours is but a first foray into new territory we hope it leads to new frontiers. Indeed, there are several exploratory avenues that come to mind:
\begin{itemize}
\item First, we would like to add other and more complex problem domains, hoping to reinforce the promising results presented herein and identify areas where the SAFE concept is not only competitive but clearly advantageous. It is of particular interest to demonstrate whether SAFE can adapt itself to problems where little prior knowledge exists regarding either the objective or the best path to said objective. 

\item The coevolutionary dynamics engendered by SAFE are likely to be quite interesting and worthy of study in and of themselves. The impact of elements including elitism approach, run parameters, and convergence criteria are all worthy of future consideration.

\item \sloppy Our evolving objective function comprises two components, distance-to-objective and novelty. We could consider other components that might be added to the mix, e.g., the robot's distances to walls, its trajectory (behavior) through the environment \cite{Gomez:2009,doncieux2013behavioral}, and so forth.

\item We used a simple measure to drive objective-function evolution---genotypic novelty. Other, possibly better measures might be designed. As noted in Section~\ref{sec:safe} we did preliminary work on mutualistic fitness scores for objective functions, work which---though it did not pan out---we plan to continue.

\item Explore the concept of whether having an objective function that changes over the course of evolutionary search may itself be an important aspect of evolvability in certain deceptive domains. 
\end{itemize}

Considering the ``coadaptations of organic beings to each other'' \cite{Darwin:1859}, is it not natural to view the objective function as a constantly shifting, evolving entity?  We speculate that leveraging a combination of open-endedness and coevolutionary search may offer a path to solving a variety of future problems where traditional approaches fail, characterized by deceptive landscapes, high dimensionality, and a lack of prior knowledge. 


\section*{Acknowledgements}
This work was supported by National Institutes of Health (USA) grants AI116794, LM010098, and LM012601.


\bibliographystyle{splncs03}
\bibliography{safe}

\end{document}